\documentclass{article}
\usepackage{spconf,amsmath,graphicx}

\usepackage{times}
\usepackage{epsfig}
\usepackage{amsmath}
\usepackage{amssymb}

\usepackage{multirow}
\usepackage{booktabs}
\usepackage{makecell}
\usepackage{setspace}
\usepackage{bm}

\newcommand{\ra}{\rightarrow}

\title{Optical Flow Estimation via Motion Feature Recovery}
\name{Yang Jiao $^{\dagger, \ddagger}$ \sthanks{This work is supported by the China Scholarship Council.}, Guangming Shi $^{\dagger}$, Trac D. Tran $^{\ddagger}$}
\address{$^{\dagger}$ Xidian University, Xi'an, Shaanxi, China\\
$^{\ddagger}$ Johns Hopkins University, Baltimore, Maryland, USA}
\begin{document}
%
\maketitle
\begin{abstract}
Optical flow estimation with occlusion or large displacement is a problematic challenge due to the lost of corresponding pixels between consecutive frames. In this paper, we discover that the lost information is related to a large quantity of motion features (more than $40\%$) computed from the popular discriminative cost-volume feature would completely vanish due to invalid sampling, leading to the low efficiency of optical flow learning. We call this phenomenon the \textit{Vanishing Cost Volume Problem}. Inspired by the fact that local motion tends to be highly consistent within a short temporal window, we propose a novel iterative Motion Feature Recovery (MFR) method to address the vanishing cost volume via modeling motion consistency across multiple frames. In each MFR iteration, invalid entries from original motion features are first determined based on the current flow. Then, an efficient network is designed to adaptively learn the motion correlation to recover invalid features for lost-information restoration. The final optical flow is then decoded from the recovered motion features. Experimental results on Sintel and KITTI show that our method achieves state-of-the-art performances. In fact, MFR currently ranks second on Sintel public website.

\end{abstract}

\begin{keywords}
CNN, Optical Flow, Cost Volume, Feature Recovery, Motion Consistency
\end{keywords}

\section{Introduction}
\label{sec:intro}

Optical flow describes the 2D displacement from the current frame $I_{t}$ to the next frame $I_{t+1}$, providing the essential motion clues for numerous practical applications such as autonomous driving \cite{Menze} and action recognition \cite{ActionRecognition}. To learn the pixel-wise mapping from image domain to optical flow domain, the critical key is to accurately extract motion features between two neighboring video frames.

In existing methods, FlowNet \cite{FlowNet} is the pioneering work to employ convolutional neural network (CNN) for motion feature extraction. FlowNet2 \cite{FlowNet2} refines and improves the results by stacking additional CNN blocks. To discriminatively describe the motion feature, the concept of \textit{Cost Volume} is widely exploited in many recent techniques \cite{PWC-Net, ScopeFlow, LiteFlowNet, MaskFlowNet, UnRigidFlow, RAFT} since it captures the high level matching cost for associating a pixel with its neighbors. For example, PWC-Net \cite{PWC-Net} constructs a partial cost volume to evaluate the matching similarity for warped features, whereas a 4D all-pair cost volume with recurrent units is proposed in RAFT \cite{RAFT}, leading to new state-of-the-art (SOTA) benchmark results.

\begin{figure}[htbp]
\begin{center}
    \includegraphics[width=0.75\linewidth]{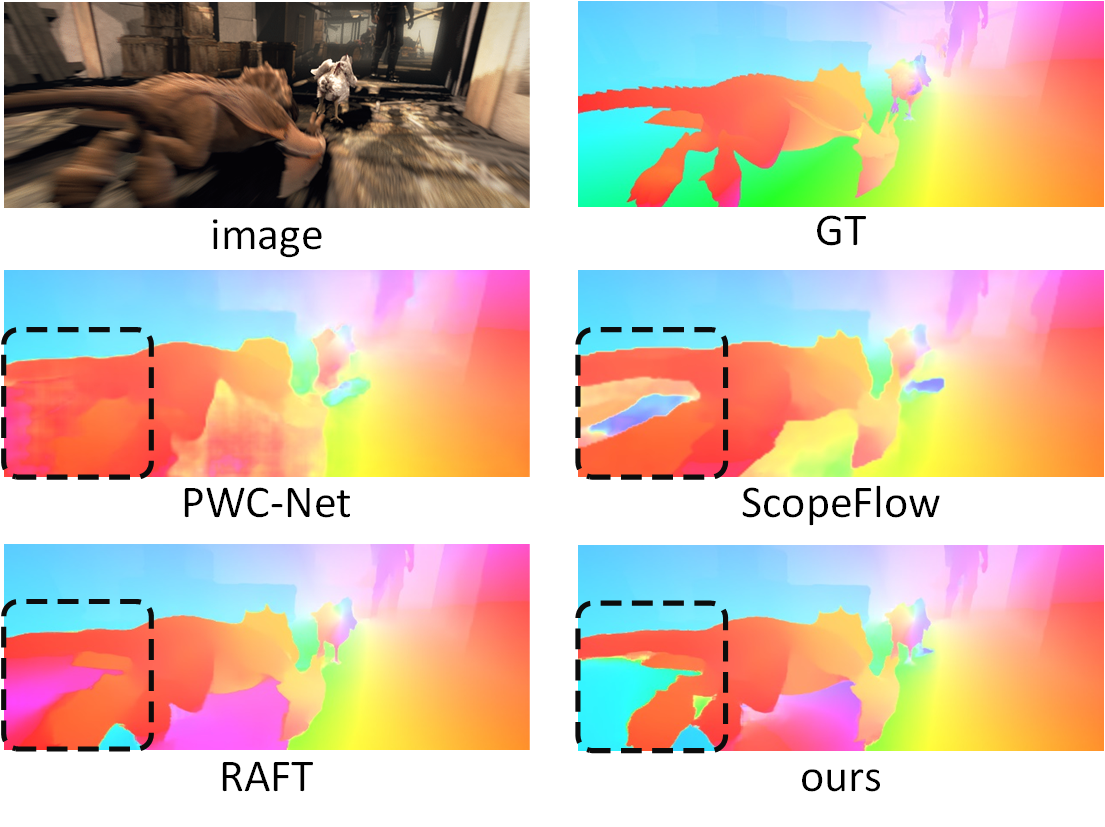}
\end{center}
\vspace{-5mm}
    \caption{Existing methods predict inaccurate optical flow with large displacement. Our method could generate better results via motion feature recovering.}
\label{fig:IntroCompare}
\end{figure}

Although satisfying results can be achieved, existing methods are still easily prone to failure when the scene is occluded or there exists large motion displacement as shown in Figure \ref{fig:IntroCompare}. To address this challenge, we re-investigate the feature matching process in optical flow and we find that in the situation of occlusion or fast motion, a high percentage of motion features (more than $40\%$) sampled from cost volume vanishes to all zeros. This in turn will definitely corrupt the feature matching process during the optical flow learning stage, leading to the low efficiency when learning with incomplete motion features.

To address this the \textit{Vanishing Cost Volume Problem}, we rely on the fact that local motion is highly consistent within a short time-frame $\Delta t$. Our key insight to solving the Vanishing Cost Volume Problem is that the missing motion features from time $t$ to $t+1$ can be inferred from the motion history by modeling correlation along a series of consecutive multiple frames. To this end, we propose a novel Motion Feature Recovery method to adaptively learn the inherent motion relation across frames. MFR is designed in an iteration fashion. In each iteration, we first extract the motion feature $M$ by CNN blocks, and then determine the invalid entries of $M$ based on the coarse motion flow. Next, an efficient network is applied to adaptively model the motion correlation for recovering the vanished motion feature. Lastly, better optical flow can be decoded from the recovered feature with rich motion information via the Gated Recurrent Unit (GRU). Our original contributions can be summarized as follows.
\begin{itemize}
    \item We show that the \textit{Vanishing Cost Volume Problem} is prominent in optical flow task, and demonstrate that addressing this issue is the key to handle occlusion and large displacement.

    \item We propose a solution to the above problem by recovering invalid motion features from learning motion consistency across frames.

    \item We design an iterative MFR structure to adaptively update motion features for better optical flow estimation.
\end{itemize}

Extensive experiments conducted on Sintel \cite{Sintel} and KITTI \cite{KITTI_2015} data sets show that MFR achieves state-of-the-art performances when benchmarking against existing optical flow estimation approaches.

\section{Method}
\label{sec:method}

We first introduce the Vanishing Cost Volume Problem in Section \ref{sec:2.1}, then propose MFR method in Section \ref{sec:2.2}. Overall network structure is given in Section \ref{sec:2.3}.

\subsection{Vanishing Cost Volume Problem}
\label{sec:2.1}

\begin{figure}[htbp]
\begin{center}
    \includegraphics[width=0.80\linewidth]{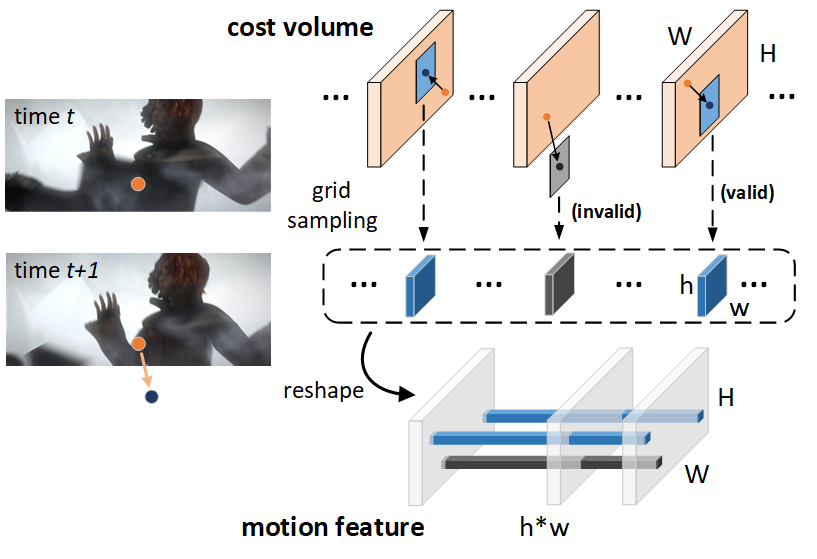}
\end{center}
    \caption{Illustration of Vanishing Cost Volume Problem at a single feature pyramid scale.}
\label{fig:CV vanishing}
\end{figure}

Cost volume encodes the feature pattern similarity between two features $g_\theta(I_t) \in \mathbb{R} ^{D \times H \times W}$ and $g_\theta(I_{t+1}) \in \mathbb{R}^{D \times H \times W}$, where $g_\theta$ is the feature extractor. A 3D cost volume $C \in \mathbb{R}^{(H \times W) \times H \times W}$ can be formulated as:
\begin{equation}
\label{eq:cost volume}
C(kW+l, i, j) = \sum_{d=0}^{D-1} g_\theta(I_t)(d, i, j) \cdot g_\theta(I_{t+1})(d, k, l),
\end{equation}
where $(d, i, j)$ and $(d, k, l)$ indicate the element position in the current feature map $g_\theta(I_t)$ and the next $g_\theta(I_{t+1})$. Given a cost volume $C$, the motion feature $M$ can be extracted by grid sampling from $C$. The size of sampling grid is $h \times w$, and the center is determined by the current optical flow $F_{t \ra {t+1}}$ as illustrated in Figure \ref{fig:CV vanishing}. In this manner, the motion feature $M$ is able to store the most relevant matching points for each pixel in frame $I_t$, and the optical flow can be then efficiently decoded from $M$ via CNN blocks.

This mechanism works perfectly for most of the video scene. However, when the object is heavily occluded or moving fast with large displacement, the generated motion feature $M$ may be sampled from the area which extends beyond the cost volume boundary, resulting in the vanishing of cost volume issue. Specifically, as illustrated in Figure \ref{fig:CV vanishing}, the red pixel from time $t$ moves to the black dot at time $t+1$ with a large displacement. This leads to invalid cost volume sampling (shown in black color feature map) during motion feature generation. In other words, no valuable motion information is passed forward to the flow decoder from the vanished cost volume elements. Moreover, we also found that these invalid entries even make up a large portion ($\sim 40 \%$) in occlusion or fast moving situations, which severely harm the learning process. Therefore, recovering the invalid entries from $M$ becomes the key for successful optical flow estimation.
\begin{figure*}[htbp]
\begin{center}
    \includegraphics[width=0.75\linewidth]{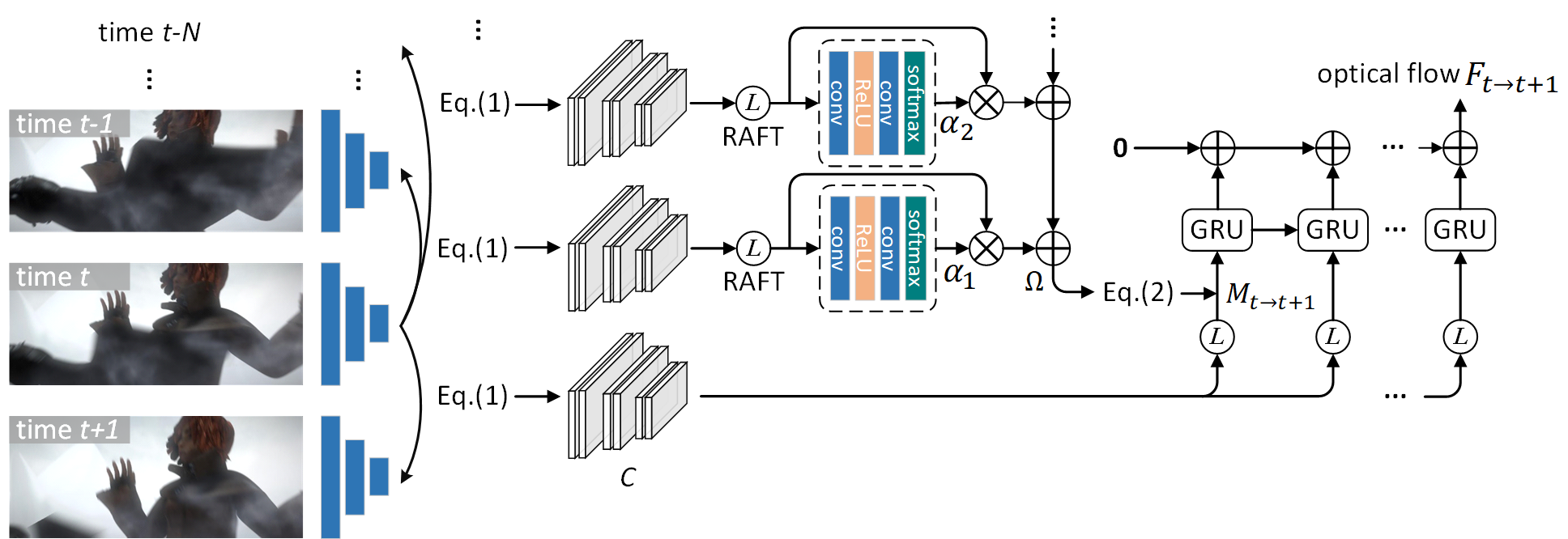}
\end{center}
    \caption{Network structure of the proposed method. We only show one iteration of GRU for simplicity.}
\label{fig:NetworkStructure}
\end{figure*}
\subsection{Motion Feature Recovery (MFR)}
\label{sec:2.2}
Occluded motion is difficult to estimated only from two neighboring frames. We recover the invalid motion feature by modeling the multi-frame motion consistency.

It is known that in a very short period of time $\Delta t$, object motion $F_{t \ra {t+1}}$ from time $t$ to $t+1$ should be highly consistent with $F_{{t-1} \ra {t}}$ from time $t-1$ to $t$. Historical optical flow $F_{{t-1} \ra t}$ provides the additional motion information to infer the current flow $F_{t \ra {t+1}}$. Therefore, the vanished elements in $M_{t \ra {t+1}}$ can be recovered by a series of $M_{{t-n} \ra t}$, where $n=\{1, ..., N\}$ is the previous time stamp. To strictly enforce that each point from previous frames could precisely match the same pixel at time $t$, we use the inverse motion feature $M_{t \ra {t-n}}$ from $t$ to $t-n$ to approximate $M_{t \ra {t+1}}$ within a local patch with size $\mathcal{N}$ as follows:
\begin{equation}
\label{eq:consistency}
M_{t \ra {t+1}}^\Omega = \frac{1}{\mathcal{N}N} \sum_{\mathcal{N}} \sum_{n=1}^{N} \alpha_{n} \cdot M_{t \ra {t-n}}^\Omega,
\end{equation}
where $\Omega$ is the support of zero entries in $M_{t \ra {t+1}}$, and the non-zero values in $M_{t \ra {t+1}}$ is unchanged. Also, $\alpha_{n} \in \mathbb{R}^{\mathcal{N} \times H \times W}$ is the coefficient matrix from $t$ to $t-n$.

To achieve this, we design a simple but effective network to adaptively learn the coefficient $\alpha_n$. Specifically, for each $n$, motion feature $M_{t \ra {t-n}}$ is fed into a two-layer CNN connected by a ReLU activation. Then, a SoftMax layer is applied for spatial normalization.

\subsection{Overall Structure}
\label{sec:2.3}

We use RAFT \cite{RAFT} as the backbone, and construct our model in an iterative fashion. Figure \ref{fig:NetworkStructure} depicts the structure of one GRU iteration.

In the figure, frames from $t-N$ to $t+1$ are first represented by a shared CNN feature extractor. Then \ref{eq:cost volume} computes the cost volumes for all pairs in different pyramid level. We perform RAFT for frame pair $\{I_{t}, I_{t-n}\}$ for $n=1$ to $N$, and compute the historical motion feature $M_{t \ra {t-n}}$ via the table-lookup operation $L$ which is defined in \cite{RAFT}. Coefficients $\alpha_n$ in \ref{eq:consistency} are learned from $M_{t \ra {t-n}}$ via a two-layer CNN followed by SoftMax normalization, and the invalid features in $M_{t \ra {t+1}}$ can be recovered by \ref{eq:consistency}. Finally, GRU decodes the optical flow from the reconstructed $M_{t \ra {t+1}}$. To optimize the network, we follow the same strategy as in the \cite{RAFT} framework -- minimizing the $l$-2 loss between the predicted flow and labels at the end of each GRU.

\section{Experiment}
\label{sec:exp}

\subsection{Implementation Details}
As in \cite{PWC-Net, ScopeFlow, RAFT}, we pre-train our model on FlyingChair (F) \cite{FlyingChairs} $\ra$ FlyingThings (T) \cite{FlyingThings} for 100K iterations each with batch size 12 $\ra$ 6, and then use two splits for evaluation. In the C+T+S/K split, we fine-tune the model on Sintel (S) \cite{Sintel} for the final 100k iterations or on KITTI-2015 (K) \cite{KITTI_2015} for 50k. In the C+T+S+K+H split, the combination of Sintel, KIITI and HD1D (C+K+H) is used for fine-tuning with 100k (for Sintel) and 50K (for KITTI) iterations.

All modules in the training are randomly initialized. Images from Sintel are fixed to $368 \times 768$ size, and $288 \times 960$ for KITTI. History frame number $N$ is set to 2, meaning that 4 frames from $I_{t-2}$ to $I_{t+1}$ are used to predict flow $F_{t \ra {t+1}}$. Window size $\mathcal{N}=1$. AdamW is adopted as the optimizer, and the weight decay is set to $1e$-4. All experiments are implemented in PyTorch and trained on two Tesla P40 GPUs.

\begin{table*}[htbp]
\small
\begin{spacing}{1.0}
\begin{center}
    \caption{Quantitative comparison of optical flow with EPE and Fl employed for evaluation.}
\label{tab:Evaluation}
\setlength{\tabcolsep}{1.45mm}{
\begin{tabular}{p{2.5cm}<{\centering} p{3cm} | p{1.0cm}<{\centering} p{1.0cm}<{\centering} p{1.0cm}<{\centering} p{1.0cm}<{\centering} | p{1.0cm}<{\centering} p{1.0cm}<{\centering} p{1.0cm}<{\centering}}

    \bottomrule[1.5pt]
    \multirow{2}{2cm}{\centering Training Data} & \multirow{2}{2cm}{\centering Method} & \multicolumn{2}{c}{Sintel (train)} & \multicolumn{2}{c|}{Sintel (test)} & \multicolumn{2}{c}{KITTI (train)} & \multicolumn{1}{c}{KITTI (test)}\\
    & & Clean & Final & Clean & Final & EPE-all & Fl-all & Fl-all \\

    \hline

    \multirow{6}{2cm}{\centering C+T+S/K} &
      FlowNet2  \cite{FlowNet2}  & (1.45) & (2.01) & 4.16 & 5.74 & (2.30) & (6.8) & 11.48 \\
    & HD3       \cite{HD3}       & (1.87) & (1.17) & 4.79 & 4.67 & (1.31) & (4.1) & 6.55 \\
    & IRR-PWC   \cite{IRR}       & (1.92) & (2.51) & 3.84 & 4.58 & (1.63) & (5.3) & 7.65 \\
    & VCN       \cite{VCN}       & (1.66) & (2.24) & 2.81 & 4.40 & (1.16) & (4.1) & 6.30 \\
    & ScopeFlow \cite{ScopeFlow} & -      & -      & 3.59 & 4.10 & -      & -     & 6.82 \\
    & RAFT      \cite{RAFT}      & (0.77) & (1.20) & 2.08 & 3.41 & (0.64) & (1.5) & 5.27 \\
    & Ours                       & \textbf{(0.65)} & \textbf{(1.01)} & \textbf{2.01} & \textbf{3.29} & \textbf{(0.59)} & \textbf{(1.3)} & \textbf{5.17} \\

    \hline

    \multirow{6}{2cm}{\centering C+T+S+K+H} &
      LiteFlowNet2  \cite{LiteFlowNet2} & (1.30) & (1.62) & 3.48 & 4.69 & (1.47) & (4.8) & 7.74 \\
    & PWC-Net+      \cite{PWC-Net+}     & (1.71) & (2.34) & 3.45 & 4.60 & (1.50) & (5.3) & 7.72 \\
    & MaskFlowNet   \cite{MaskFlowNet}  & -      & -      & 2.52 & 4.17 & -      & -     & 6.10 \\
    & RAFT          \cite{RAFT}         & (0.76) & (1.22) & 1.94 & 3.18 & (0.63) & (1.5) & 5.10 \\
    & RAFT-warm     \cite{RAFT}         & (0.77) & (1.27) & 1.61 & 2.86 & -      & -     & -    \\
    & Ours                              & \textbf{(0.64)} & \textbf{(1.04)} & \textbf{1.55} & \textbf{2.80} & \textbf{(0.54)} & \textbf{(1.1)} & \textbf{5.03} \\

    \toprule[1.5pt]

\end{tabular}}
\end{center}
\end{spacing}
\vspace{-6mm}
\end{table*}

\subsection{Evaluation}
Optical flow comparisons conducted on Sintel and KITTI datasets are summarized in Table \ref{tab:Evaluation}.  Averaged end-point-error (EPE) and percentage of optical flow outliers (Fl) are used for evaluation. Values in the brackets indicate that training and testing are on the same dataset. Our method achieves the best results on both Sintel and KITTI.

\begin{figure*}[htbp]
\begin{center}
    \includegraphics[width=0.90\linewidth]{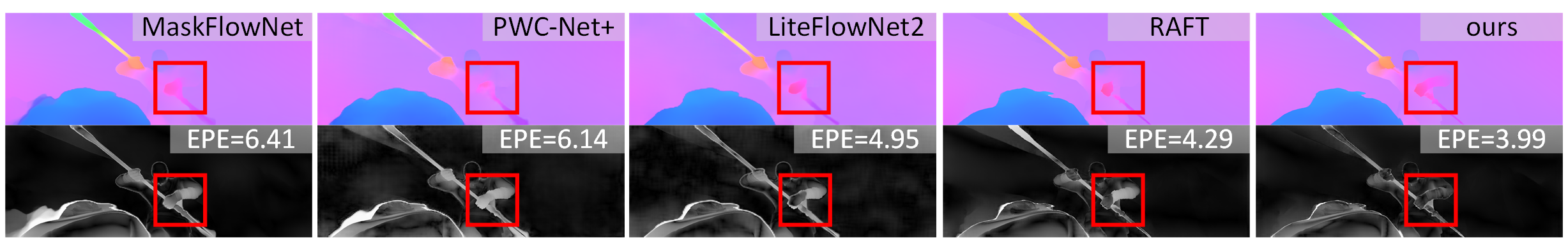}
\end{center}
\vspace{-4mm}
    \caption{Qualitative comparisons on Sintel dataset. Our method achieves better visual result and lower error.}
\label{fig:FlowCompare}
\end{figure*}

\textbf{Sintel.} This data set contains two different passes: Clean and Final pass. The former only contains basic objects while the latter is rendered with more diverse factors, such as climate change, motion blur, etc.. Our method outperforms the existing methods for both Clean and Final passes. With C+T+S training, we achieve EPE of 2.01 and 3.29, compared with SOTA 2.08 and 3.41 from RAFT \cite{RAFT}. Training with more data from KITTI (K) and HD1K (H) improves the generalization of the model and significantly reduces the testing error from 3.29 to 2.81. Our results demonstrate that recovering motion feature from multiple frames are indeed effective. Several qualitative comparisons are illustrated in Figure \ref{fig:FlowCompare}.

\textbf{KITTI.} KITTI is more challenging than Sintel due to the lack of training samples and sparse ground truth. The proposed method consistently achieves SOTA results for both training and testing set. Compared with training set, the performance gain on testing set is marginal. We hypothesize that this is due to having access to only 200 samples in training for KITTI, which may lead to over-fitting and in turn may limit the model generalization ability.

\subsection{Effectiveness of Motion Feature Recovery}
\label{sec:3.4}

\begin{figure}[ht]
\begin{center}
    \includegraphics[width=1.0\linewidth]{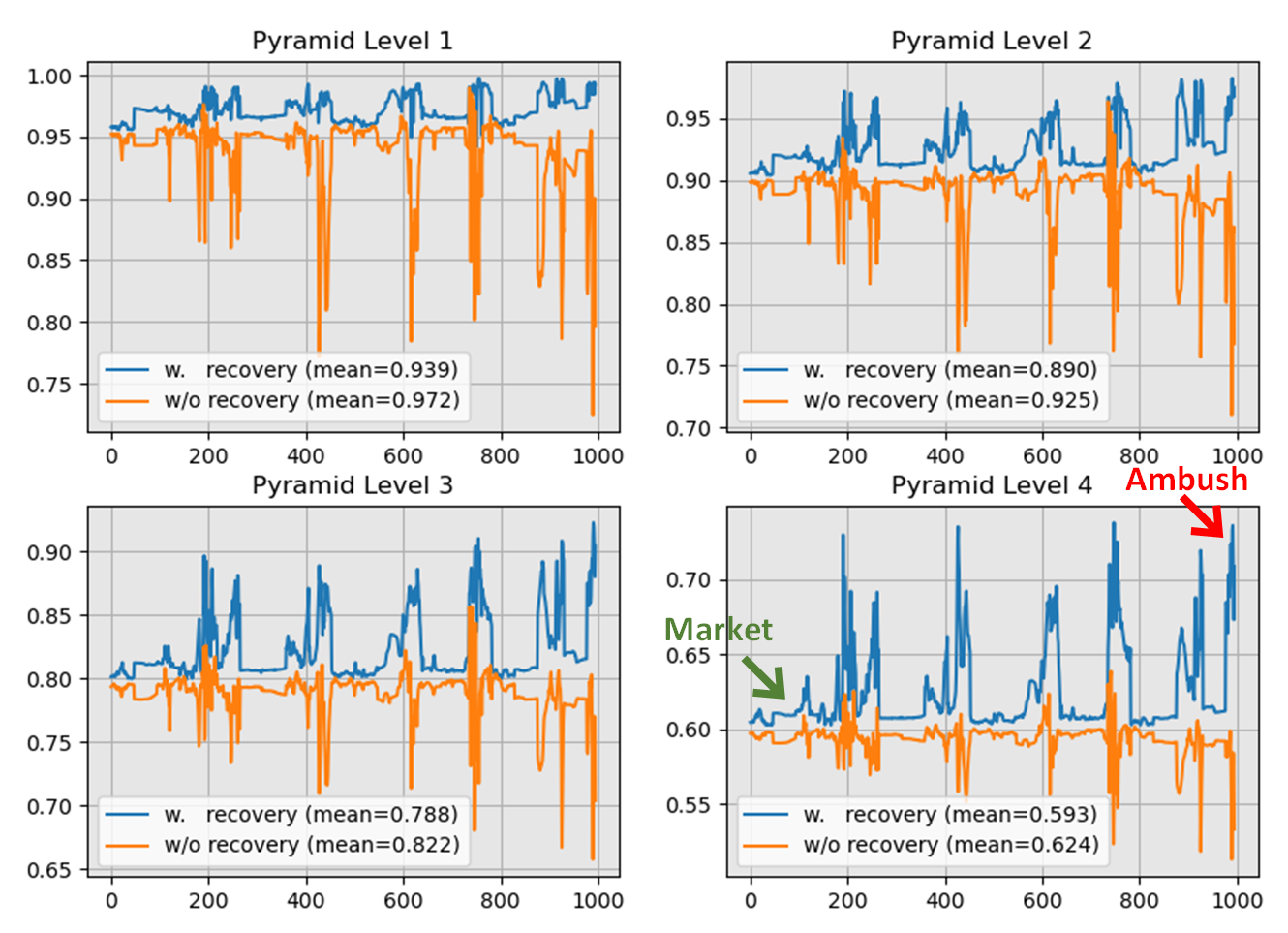}
\end{center}
\vspace{-5mm}
    \caption{Comparisons of non-zero element ratio in motion feature $M_{t \ra {t+1}}$. Our method (w. recovery) outputs higher NZR values for each pyramid level.}
\vspace{-5mm}
\label{fig:NZR}
\end{figure}

Non-zero element ratio (NZR) in motion feature $M_{t \ra {t+1}}$ indicates the amount of valid motion information that can be used for flow estimation. We compare the NZR with and without using the proposed MFR on Sintel training set in Figure \ref{fig:NZR}. The horizontal axis represents the sample index.

Figure \ref{fig:NZR} illustrates the NZR from four pyramid levels. For each level, our method (w. recovery) consistently generates higher NZR values (blue curve) than without using MFR (red curve), demonstrating the effectiveness of the proposed motion feature recovery strategy. It also can be observed that the averaged NZR value from higher pyramid level (0.593 in level-4) is always lower than the lower level (0.939 in level-1). This is because the fixed motion vectors in higher level represents larger object displacements, which are more likely to extend beyond the image boundary.

Motion feature recovery in our model significantly improves the NZR for more challenging cases. For example, NZR from "Ambush" subset can be improved from a very low value of 0.51 to 0.73 (a $30.1\%$ gain) after motion recovery whereas we can only improve NZR from 0.58 to 0.62 from the easier "Market" subset (a modest $6.4\%$ gain).

\subsection{Model Analysis}
The total number of learnable parameters in the proposed model is 6.0M, which is less than most of the existing methods: FlowNet2 (162M), PWC-Net+ (9.4M), IRR-PWC (6.4M) and VCN (6.2M). Due to the additional network structure for recovering invalid motion feature, our model adds 0.7M more parameters comparing to RAFT.

\section{Conclusion}
\label{sec:conclusion}
In summary, we propose a novel MFR method for optical flow estimation via adaptive learning of the inherent correlation between consecutive multiple frames. MFR recovers invalid motion features during the matching process, providing a solution of the Vanishing Cost Volume Problem caused by either occlusion or large displacement. Extensive experiments on different optical flow benchmarks produce SOTA results. In our future work, higher order approximation and feature consistency will be explored for better estimation.

\bibliographystyle{IEEEbib}
\bibliography{paperMFR}

\end{document}